\documentclass[10pt,twocolumn,letterpaper]{article}
\pdfoutput=1

\usepackage{wacv}
\usepackage{times}

\usepackage{amsmath}
\usepackage{amssymb}
\usepackage{booktabs}
\usepackage{color}
\usepackage{float}
\usepackage{graphicx}
\usepackage{paralist}
\usepackage{tabularx}
\usepackage{url}
\usepackage{xspace}
\usepackage{balance}
\usepackage[pagebackref=false,breaklinks=true,colorlinks,bookmarks=false,urlcolor=blue,linkcolor=blue,citecolor=blue,pdftitle={Spatio-Temporal Covariance Descriptors for Action and Gesture Recognition},pdfauthor={Andres Sanin, Conrad Sanderson, Mehrtash T. Harandi, Brian C. Lovell}]{hyperref}

\newcommand{\alg}[1]{\mbox{Algorithm~\ref{#1}}}
\newcommand{\fig}[1]{\mbox{Fig.~\ref{#1}}}
\newcommand{\tab}[1]{\mbox{Table~\ref{#1}}}

\newcommand{\rh}[1]{\rotatebox{66}{\bf #1}}
\newcolumntype{x}{>{\centering}X}

\newcommand{\eqsize}{\footnotesize}
\newcommand{\cov}{{Cov3D}}
\newcommand{\wrlpp}{WRLPP}
\newcommand{\vect}[1]{{\boldsymbol{#1}}}
\newcommand{\mat}[1]{{\boldsymbol{#1}}}
\newcommand{\tens}[1]{#1}

\floatstyle{ruled}
\newfloat{algorithm}{!tb}{loa}
\floatname{algorithm}{Algorithm}

\wacvfinalcopy 

\begin{document}

\title{Spatio-Temporal Covariance Descriptors for Action and Gesture Recognition}

\author
  {
  {\it Andres~Sanin, Conrad~Sanderson, Mehrtash~T.~Harandi, Brian~C.~Lovell}\\
  ~\\
  NICTA, PO Box 6020, St Lucia, QLD 4067, Australia~%
  \thanks
    {%
    {\bf Published~in:} IEEE Workshop on Applications of Computer \mbox{Vision}, pp.~103--110, 2013.
    \href{\bf http://dx.doi.org/10.1109/WACV.2013.6475006}{http://dx.doi.org/10.1109/WACV.2013.6475006}.
    {\tiny \mbox{\bf Acknowledgements:} NICTA is funded by the Australian Government via the {\it Department of Broadband, Communications and the Digital Economy}, and the Australian Research Council through the {\it ICT Centre of Excellence} program.}%
    }
  \\
  University of Queensland, School of ITEE, QLD 4072, Australia%
  }

\maketitle
\thispagestyle{empty}

\begin{abstract}
\vspace{-2ex}

\noindent
We propose a new action and gesture recognition method based on spatio-temporal covariance descriptors
and a weighted Riemannian locality preserving projection approach
that takes into account the curved space formed by the descriptors.
The weighted projection is then exploited during boosting to create
a final multiclass classification algorithm that employs the most useful spatio-temporal regions.
We also show how the descriptors can be computed quickly through the use of integral video representations.
Experiments on the UCF sport, CK+ facial expression and Cambridge hand gesture datasets
indicate superior performance of the proposed method
compared to several recent state-of-the-art techniques.
The proposed method is robust and does not require additional processing of the videos,
such as foreground detection, interest-point detection or tracking. 




\end{abstract}
\vspace{-1ex}
\section{Introduction}
\label{sec:introduction}

Video-based classification plays a key role in human motion analysis fields such as action and gesture recognition.
Both fields have shown promising applications in many areas,
including security and surveillance, content-based video analysis,
human-computer interaction and animation.
According to a recent survey on recognition of human activities~\cite{TuragaEtAl2008},
the focus has shifted to methods that do not rely on human body models,
where the information is extracted directly from the images
and hence being less dependent on reliable segmentation and tracking algorithms.
Such image representation methods can be categorised into 
global and local based approaches~\cite{Poppe2010}.

Methods with global image representation encode visual information as a whole.
Ali and Shah~\cite{AliAndShah2010} extract a series of kinematic features based on optical flow.
A group of kinematic modes is found using principal component analysis.
Guo \etal~\cite{GuoEtAl2010} encode the same kinematic features using sparse representation of covariance matrices.
Several methods first divide the region of interest into a fixed spatial or temporal grid,
extract features inside each cell and then combine them into a global representation.
For example, this can be achieved using local binary patterns (LBP)~\cite{KellokumpuEtAl2008},
or histograms of oriented gradients (HOG)~\cite{ThurauAndHlavac2008}.
Global representations are sensitive to viewpoint, noise and occlusions which may lead to unreliable classification.
Furthermore, global representations depend on reliable localisation of the region of interest~\cite{Poppe2010}.

Local representations are designed to deal with the abovementioned issues by describing the visual information as a collection of patches,
usually at the cost of increased computation.
Laptev and Lindeberg~\cite{LaptevAndLindeberg2003} extract interest points using a 3D Harris corner detector
and use the points for modelling the actions.
One of the major drawbacks is the low number of interest points that are able to remain stable across an image sequence.
A common solution is to work with windowed data,
extracting salient regions which can be represented using Gabor filtering~\cite{DollarEtAl2005}.

Wang \etal~\cite{WangEtAl2009} showed that dense sampling approaches tend to perform better compared to interest point based approaches. Dense
sampling is typically done for a set of patches inside the region of interest. Features are extracted from each patch to form a descriptor. These
descriptor representations differ from grid-based global representations in that they can have an arbitrary position and size, and that the patches
are not combined to form a single representation but form a set of multiple representations.
Examples are HOG and HOF (histogram of oriented flow) descriptors~\cite{LaptevEtAl2008},
SIFT descriptors~\cite{Lowe2004},
and their respective spatio-temporal versions, HOG3D~\cite{WangEtAl2009} and 3D SIFT~\cite{ScovannerEtAl2007}.
Because of the likely large number of descriptors and/or their high dimensionality, comparing sets of
descriptors is often not straightforward.
This has led to compressed representations such as formulating sets of descriptors as bags-of-words~\cite{NieblesEtAl2008}.

In this paper we propose the use of spatio-temporal covariance descriptors for action and gesture recognition tasks.
Flat region covariance descriptors were first proposed for the task of object detection and classification in images~\cite{TuzelEtAl2008}.
Each covariance descriptor represents the features inside an image region as a normalised covariance matrix.
They have led to improved results over related descriptors such as HOG,
in terms of detection performance as well as robustness to translation and scale~\cite{TuzelEtAl2008}.
Furthermore, covariance matrices provide a low dimensional representation
which enables efficient comparison between sets of covariance descriptors.

The proposed spatio-temporal descriptors, which we name \cov,
belong to the group of symmetric positive definite matrices which do not form a vector space.
They can be formulated as a connected Riemannian manifold,
and taking into account the non-linear nature of the space of the descriptors may lead to improved classification results.
The most common approach for classification on manifolds is to first map the points
into an appropriate Euclidean representation~\cite{LinAndZha2008}
and then use traditional machine learning methods. 
A recent example of mapping is the Riemannian locality preserving projection (RLPP) technique~\cite{HarandiEtAl2012}.

The \cov~descriptors are extracted from spatio-temporal windows inside sample videos,
with the number of possible windows being very large.
As such, we use a boosting approach to search the windows to find a subset which is the most useful for classification.
We propose to extend RLPP by weighting (WRLPP), in order to take into account the weights of the training samples.
This weighted projection leads to a better representation of the neighbourhoods around the most critical training samples during each boosting iteration.
The proposed \cov~descriptors, in conjunction with the classification approach based on WRLPP boosting,
lead to a state-of-the-art method for action and gesture recognition.

We continue the paper as follows.
In Section~\ref{sec:cov3d} we describe the spatio-temporal covariance descriptors,
and use the concept of integral video to enable fast calculation inside any spatio-temporal window.
In Section~\ref{sec:method}, we first overview the concept of Riemannian manifolds formulated in the context of positive definite symmetric matrices,
and then detail the proposed boosting classification approach based on weighted Riemannian locality preserving projection.
In Section~\ref{sec:experiments}, we compare the performance of the proposed method against several recent state-of-the-art methods
on three benchmark datasets.
Concluding remarks and possible future directions are given in Section~\ref{sec:discussion}.
\begin{figure}[!b]
  \centering
  \includegraphics[width=1\columnwidth]{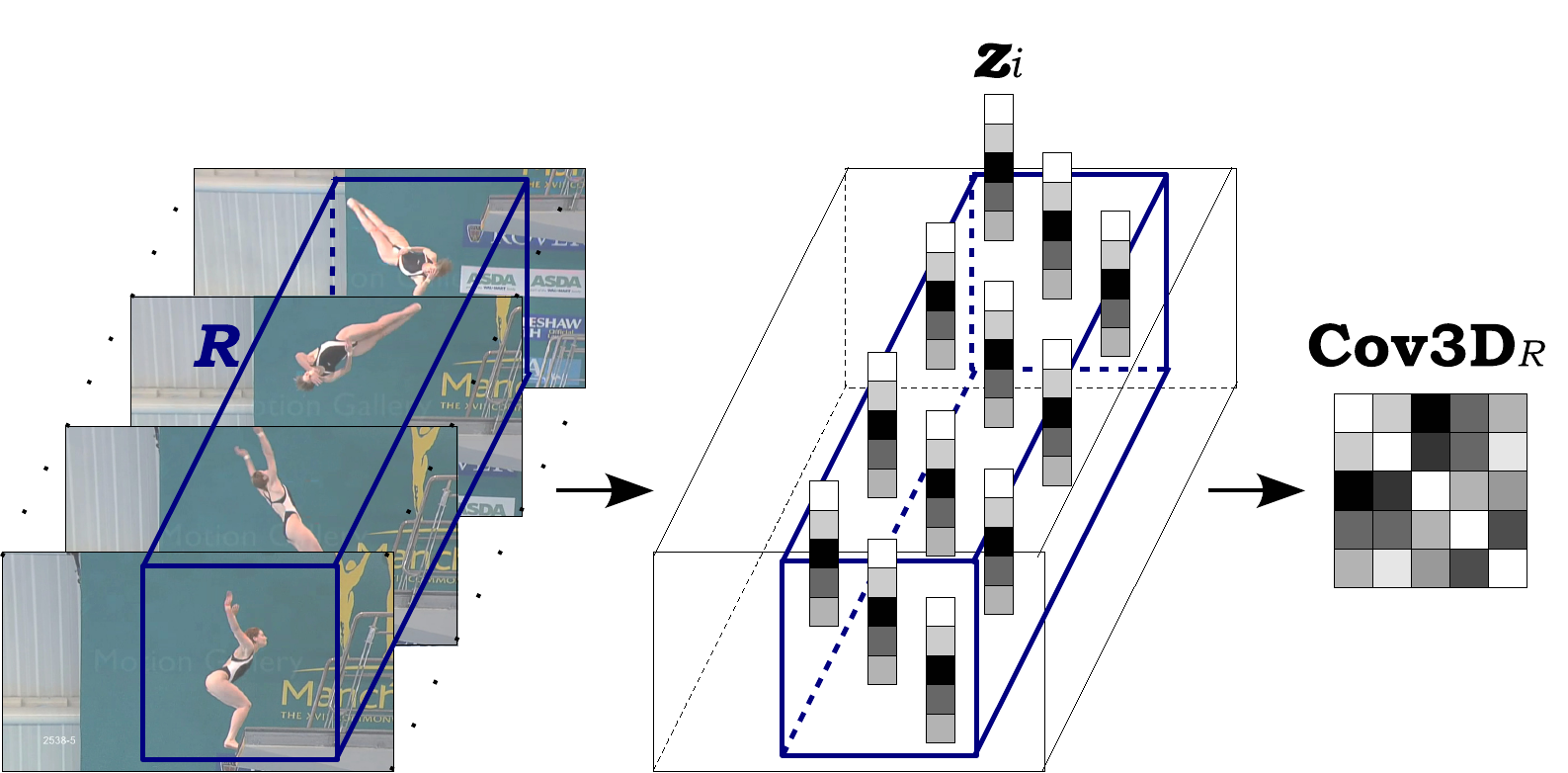}
  
  ~
  
  \caption
    {
    \small
    Conceptual demonstration for obtaining a {\cov} spatio-temporal covariance descriptor.
    A~spatio-temporal window {\eqsize $R$} is defined inside the input video.
    For each pixel in {\eqsize $R$} a feature vector {\eqsize $\vect{z}_i$} is calculated.
    The feature vectors are then used to compute the covariance matrix {\eqsize $\cov_R$}.
    }
  \label{fig:cov3d}
\end{figure}

\section{\cov~Descriptors}
\label{sec:cov3d}

In this section we first present the general form of the proposed spatio-temporal covariance descriptors (\cov),
an algorithm for their fast calculation,
and finally how they can be specialised for action and gesture recognition.
For convenience, we follow the notation in~\cite{TuzelEtAl2008}.

Let {\eqsize $V$} be the sequence of images {\eqsize ${\{I_t\}}_{t=1}^{T}$}
and {\eqsize $F$} be the {\eqsize $W  \times H \times T \times d$} dimensional feature video extracted from {\eqsize $V$}:

\vspace{-1ex}
\eqsize
\begin{equation}
  F(x,y,t) = \Phi(V,x,y,t)
  \label{eq:feature_mapping}
\end{equation}
\normalsize

\noindent
where the function {\eqsize $\Phi$} can be any mapping such as intensity, colour, gradients, or optical flow.
For a given spatio-temporal window {\eqsize $R \subset F$},
let {\eqsize ${\{\vect{z}_i\}}_{i=1}^{S}$} be the {\eqsize $d$}-dimensional feature vectors inside {\eqsize $R$}.
The region {\eqsize $R$} is represented with the
{\eqsize $d \times d$} covariance matrix of the feature vectors:

\vspace{-1ex}
\eqsize
\begin{equation}
  \cov_R = \frac{1}{S-1} \sum\nolimits_{i=1}^S(\vect{z}_i-\vect{\mu})(\vect{z}_i-\vect{\mu})^T
  \label{eq:cov}
\end{equation}
\normalsize

\noindent
where {\eqsize $\vect{\mu}$} is the mean of the points.
\fig{fig:cov3d} shows the construction of a covariance descriptor inside a spatio-temporal window.
Examples of feature vectors specific for action and gesture recognition are given in Section~\ref{sec:features}.

Representing a spatio-temporal window with a covariance matrix has several advantages:
{\bf (i)}~it is a low-dimensional representation which is independent on the size of the window,
{\bf (ii)}~the impact of noisy samples is reduced through the averaging during covariance computation,
{\bf (iii)}~it is a straightforward method of fusing correlated features.

\subsection{Fast computation}
\label{sec:integral}

Integral images are an intermediate image representation used for the fast calculation of region sums~\cite{ViolaAndJones2001}.
The concept has been extended to image sequences~\cite{KeEtAl2005},
where the integral images are stacked to form an integral video,
and can be used to compute spatio-temporal region sums in constant time.
For a video {\eqsize $V$}, its integral video {\eqsize $IV$} is defined as:

\vspace{-1ex}
\eqsize
\begin{equation}
  IV(x',y',t') = \sum\nolimits_{x \leq x'} \sum\nolimits_{y \leq y'} \sum\nolimits_{t \leq t'} V(x,y,t)
\end{equation}
\normalsize

Tuzel \etal~\cite{TuzelEtAl2008} used the integral image representations for fast calculation of flat region covariances.
Here we extend the idea for fast calculation of covariance matrices inside a spatio-temporal window using the integral video representation.
The {\eqsize $(i,j)$}-th element of the covariance matrix defined in \eqref{eq:cov} can be expressed as:

\noindent
\eqsize
\begin{equation}
  \cov_R(i,j)
  \mbox{~=~}
  \frac{1}{S \mbox{-} 1}
  \hspace{-0.5ex}
  \left[
    \hspace{0.3ex}
    \sum_{k=1}^S \hspace{-0.5ex} \vect{z}_k(i)\vect{z}_k(j) - \frac{1}{S} \sum_{k=1}^S \hspace{-0.5ex} \vect{z}_k(i) \hspace{-0.6ex} \sum_{k=1}^S \hspace{-0.5ex} \vect{z}_k(j)
    \hspace{-0.5ex}
  \right]
\end{equation}
\normalsize

\noindent
where {\small $\vect{z}_k(i)$} refers to the {\small $i$}-th element of the {\small $k$}-th vector.
To find the covariance in a given spatio-temporal window {\eqsize $R$}, we have to compute the sum of each feature dimension, {\eqsize
$\vect{z}(i)_{i=1}^{d}$}, as well as the sum of the multiplication of any two feature dimensions, {\eqsize $\vect{z}(i)\vect{z}(j)_{i,j=1 \ldots d}$}.
With {\eqsize $d$} representing the number of dimensions,
the covariance of any spatio-temporal window can be computed in {\eqsize $O(d^2)$} time, as follows.

We need to compute a total of {\eqsize $d + d^2$} integral videos.
Let {\eqsize $\tens{P}$} be the {\eqsize $W  \times H \times T \times d$} tensor of the integral videos:

\vspace{-1ex}
\eqsize
\begin{equation}
  \tens{P}(x',y',t',i) = \sum_{x \leq x'} \sum_{y \leq y'} \sum_{t \leq t'} F(x,y,t)(i)
\end{equation}
\normalsize

\noindent
where {\small $F(x,y,t)(i)$} is the {\small $i$}-th element of vector {\small $F(x,y,t)$}.
Furthermore, let {\small $\tens{Q}$} be the
 {\small $W  \times H \times T \times d \times d$} tensor of the second-order integral videos:

\vspace{-2ex}
\eqsize
\begin{equation}
  \tens{Q}(x',y',t',i,j) = \sum_{x \leq x'}\sum_{y \leq y'}\sum_{t \leq t'} F(x,y,t)(i) \cdot F(x,y,t)(j)
\end{equation}
\normalsize

\noindent
for {\small $i,j=1, \ldots, d$}.
The complexity of calculating the tensors is {\small $O(d^2WHT)$}.
The {\small $d$}-dimensional feature vector {\small $\vect{p}_{x,y,t}$}
and the {\small $d \times d$} dimensional matrix {\small $\mat{Q}_{x,y,t}$}
can be obtained from the above tensors using:

\noindent
\eqsize
\begin{eqnarray}
  \vect{p}_{x,y,t} &=& [ ~\tens{P}(x,y,t,1), ~\ldots,~ \tens{P}(x,y,t,d)~ ]^T\\
  \mat{Q}_{x,y,t} &=& \begin{pmatrix}
                  \tens{Q}(x,y,t,1,1) & \cdots & \tens{Q}(x,y,t,1,d)\\
                  \vdots       & \ddots & \vdots\\
                  \tens{Q}(x,y,t,d,1) & \cdots & \tens{Q}(x,y,t,d,d)
                \end{pmatrix}
\end{eqnarray}
\normalsize

\begin{figure}[!b]
  \vspace{-3ex}
  \centering
  \includegraphics[width=0.85\columnwidth]{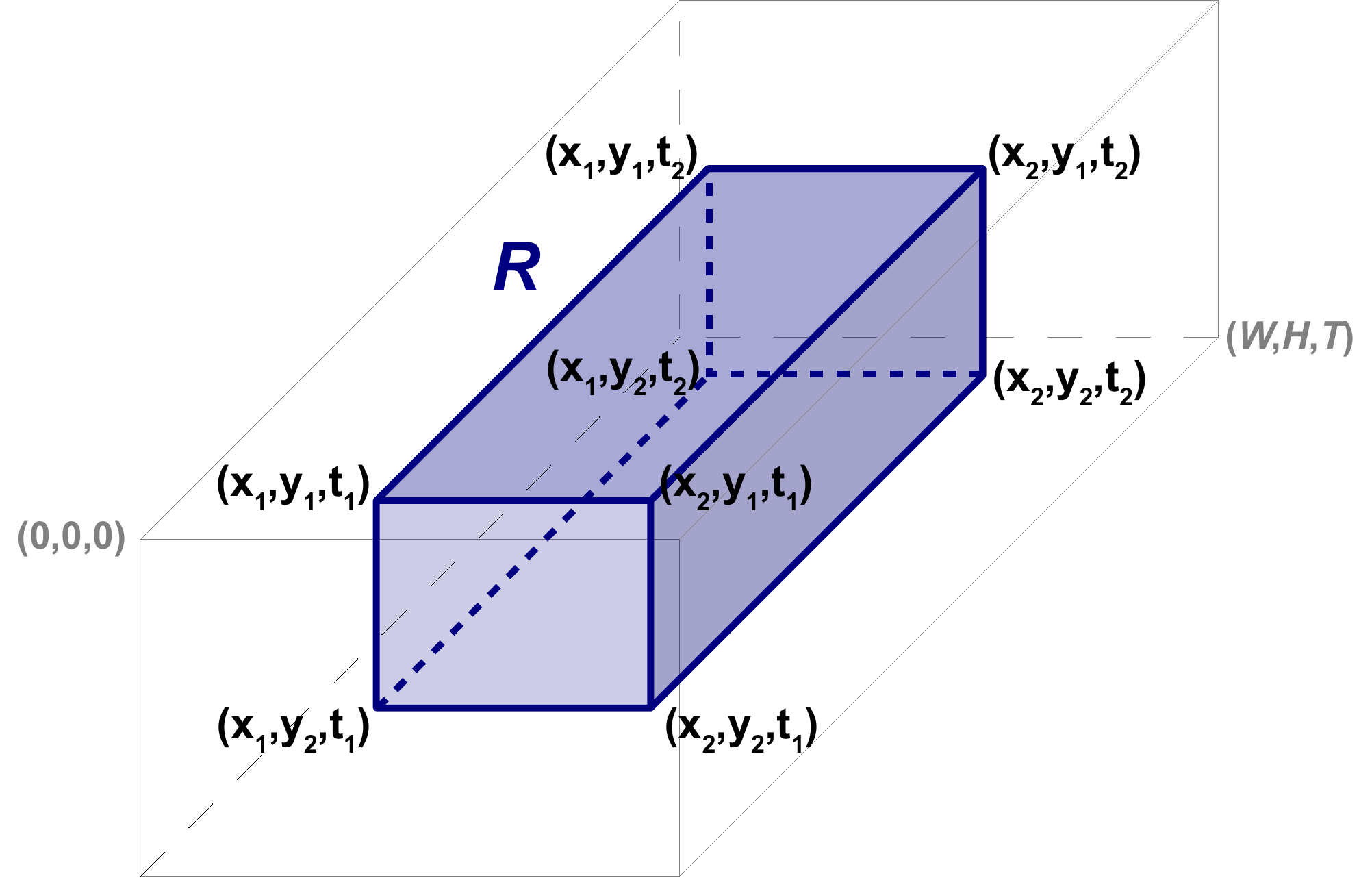}
  \caption
    {
    \small
    Integral feature video.
    The spatio-temporal window {\eqsize $R$} is bounded by {\eqsize $(x_1,y_1,t_1)$} and {\eqsize $(x_2,y_2,t_2)$}.
    Each point in {\eqsize $R$} is a {\eqsize $d$} dimensional vector,
    where {\eqsize $d$} is the number of features.
    }
  \label{fig:integral}
\end{figure}

Let {\eqsize $R(x_1,y_1,t_1;~ x_2,y_2,t_2)$}
be the spatio-temporal window of points
{\eqsize $\{(x,y,t) | x_1 \leq x \leq x_2, y_1 \leq y \leq y_2, t_1 \leq t \leq t_2\}$},
as shown in \fig{fig:integral}.
The covariance of the spatio-temporal window bounded by {\eqsize $(0,0,0)$} and {\eqsize $(x,y,t)$} is:

\vspace{-2ex}
\eqsize
\begin{equation}
  \cov_{R(0,0,0;~ x,y,t)} = \frac{1}{S-1} \left[ \mat{Q}_{x,y,t} - \frac{1}{S} \vect{p}_{x,y,t} ~ \vect{p}_{x,y,t}^T \right]
\end{equation}
\normalsize

\noindent
where {\eqsize $S = x \cdot y \cdot t$}.
Similarly, after a few rearrangements,
the covariance of the region {\eqsize $R(x_1,y_1,t_1;x_2,y_2,t_2)$} can be computed as:

\noindent
\eqsize
\begin{multline}
  \cov_{R(x_1,y_1,t_1;~ x_2,y_2,t_2)} =\\
    \frac{1}{S-1} \left[ \widehat{\mat{Q}}_{x_2,y_2} + \widehat{\mat{Q}}_{x_1-1,y_1-1} - \widehat{\mat{Q}}_{x_2,y_1-1} - \widehat{\mat{Q}}_{x_1-1,y_2} \right.\\
                         - \frac{1}{S} \left( \widehat{\vect{p}}_{x_2,y_2} + \widehat{\vect{p}}_{x_1-1,y_1-1} - \widehat{\vect{p}}_{x_2,y_1-1} - \widehat{\vect{p}}_{x_1-1,y_2} \right)\\
                  \left. \left( \widehat{\vect{p}}_{x_2,y_2} + \widehat{\vect{p}}_{x_1-1,y_1-1} - \widehat{\vect{p}}_{x_2,y_1-1} - \widehat{\vect{p}}_{x_1-1,y_2} \right)^T \right]
\end{multline}
\normalsize

\noindent
where
\vspace{-3ex}

\noindent
\eqsize
\begin{eqnarray}
  \widehat{\vect{p}}_{x,y} &=& \vect{p}_{x,y,t_2} - \vect{p}_{x,y,t_1}\\
  \widehat{\mat{Q}}_{x,y} &=& \mat{Q}_{x,y,t_2} - \mat{Q}_{x,y,t_1}
\end{eqnarray}
\normalsize

\noindent
and {\eqsize $S = (x_2 - x_1 + 1) \cdot (y_2 - y_1 + 1) \cdot (t_2 - t_1 + 1)$}.

%
%
%

\subsection{Features and regions}
\label{sec:features}
\vspace{-1ex}

Commonly used features for action and gesture recognition include intensity gradients and optical flow.
Previous studies have shown the benefit of combining both types of features~\cite{DollarEtAl2005,WangEtAl2009}.
We define the feature mapping {\eqsize $\Phi$},
present in \eqref{eq:feature_mapping},
as the following combination of gradient and optical-flow based features,
extracted from pixel location {\eqsize $(x,y,t)$}:

\vspace{-1ex}
\eqsize
\begin{equation}
  \Phi(V,x,y,t) = [ ~ x ~~ y ~~ t ~~ g ~~ o ~ ]^T
\end{equation}
\normalsize

\noindent
where
\vspace{-2ex}

\noindent
\eqsize
\begin{eqnarray}
\label{eq:gradients}
  g &=& \left[ ~ |I_x| ~~ |I_y| ~~ |I_{xx}| ~~ |I_{yy}| ~~ \sqrt{I_x^2 + I_y^2} ~~ \arctan\frac{|I_y|}{|I_x|} ~ \right]\\
\label{eq:flow}
  o &=& \left[ ~ u ~~ v ~~ \frac{\partial u}{\partial t} ~ \frac{\partial v}{\partial t} ~~
             \left( \frac{\partial u}{\partial x} + \frac{\partial v}{\partial y} \right) ~~
             \left( \frac{\partial v}{\partial x} - \frac{\partial u}{\partial y} \right) ~ \right]
\end{eqnarray}
\normalsize

The first four gradient based features in \eqref{eq:gradients} represent the first and second order intensity gradients at pixel location {\eqsize
$(x,y)$}. The last two gradient based features correspond to the gradient magnitude and gradient orientation. The optical-flow based features in
\eqref{eq:flow} represent, in order: the horizontal and vertical components of the flow vector,
the first order derivatives of the flow components with respect to {\eqsize $t$},
and the spatial divergence and vorticity of the flow field as defined in~\cite{AliAndShah2010}.
Each descriptor is hence a {\eqsize $15 \times 15$} matrix, as {\eqsize $\Phi(V,x,y,t)$} has {\eqsize $15$} dimensions.

\begin{figure}[!b]
  \centering
  \begin{minipage}{1\columnwidth}
    \begin{minipage}{0.4\columnwidth}
      \includegraphics[width=\columnwidth]{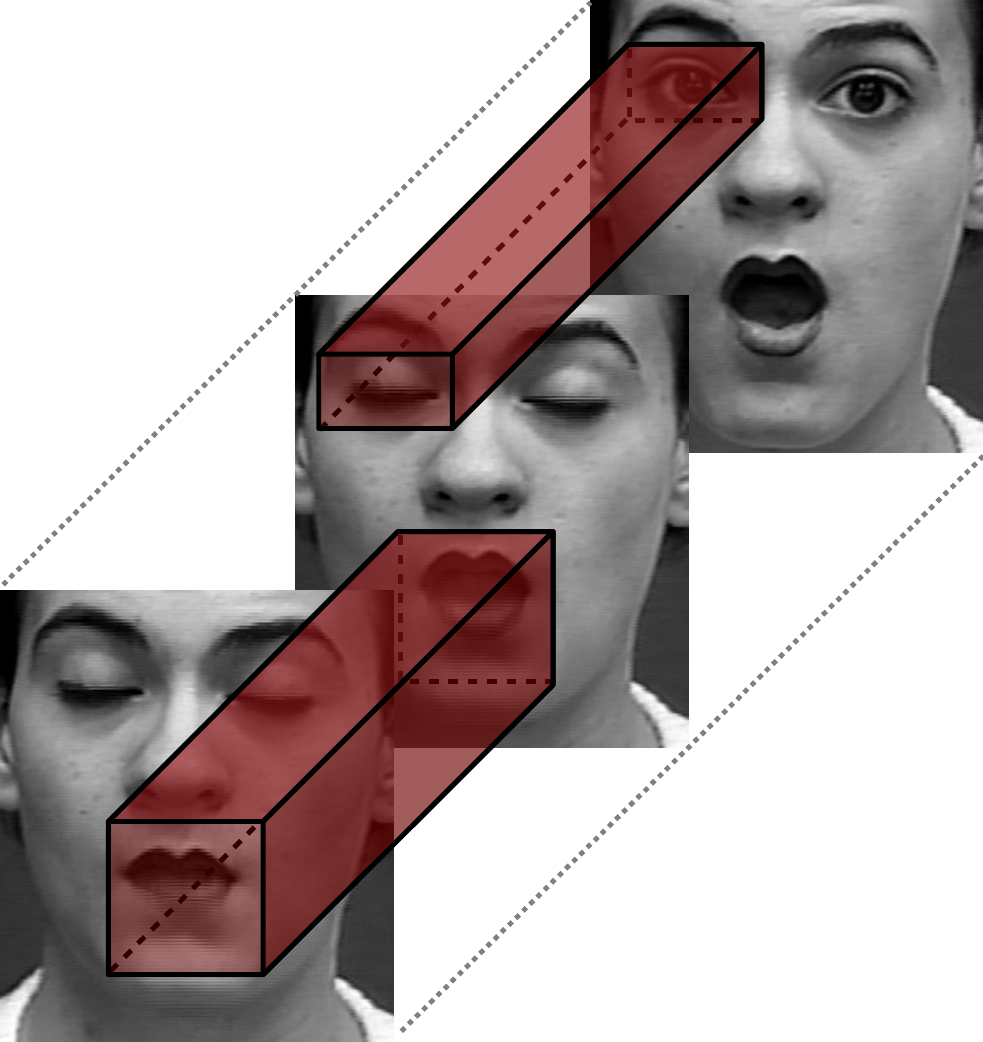}
    \end{minipage}
    \hfill
    \begin{minipage}{0.55\columnwidth}
      \caption
        {
        \small
        Two examples of \cov~windows that, together, can be useful for the recognition of face expressions.
        }
      \label{fig:stwins}
      ~
    \end{minipage}
  \end{minipage}
\end{figure}

For reliable recognition, several regions (and hence several descriptors) are typically used.
\fig{fig:stwins} shows the spatio-temporal windows of two descriptors which can be used for recognition of face expressions.
With the defined mapping, the input video {\small $V$} is mapped to {\small $F$}, a {\small $15$}-dimensional feature video.
Since the cardinality of the set of spatio-temporal windows {\eqsize $\{ R \subset F \}$} is very large,
we only consider windows of a minimum size and increment their location and size by a minimum interval value.
Further specifics on the windows used in the experiments are given in Section~\ref{sec:experiments}.

Following~\cite{TuzelEtAl2008}, each covariance descriptor {\eqsize $\cov_R$},
is normalised with respect to the covariance descriptor of the region containing the full feature video, {\eqsize $\cov_F$},
to improve the robustness against illumination variations:

\noindent
\eqsize
\begin{equation}
  \widehat{\cov}_R = \operatorname{diag}(\cov_F)^{ \mbox{-} \frac{1}{2}} ~ \cov_R ~ \operatorname{diag}(\cov_F)^{ \mbox{-} \frac{1}{2}}
\end{equation}
\normalsize

\noindent
where {\small $\operatorname{diag}(\cov_F)$} is equal to {\small $\cov_F$} at the diagonal entries and the rest is set to zero.
\section{Classification of Actions and Gestures}
\label{sec:method}

The \cov~descriptors are symmetric positive definite matrices of size \mbox{\eqsize $d \times d$},
which can be formulated as a connected Riemannian manifold ({\eqsize $Sym_d^+$})~\cite{harandi_eccv_2012}.
In this section we first briefly overview Riemannian manifolds,
followed by describing the proposed weighted Riemannian locality preserving projection (WRLPP)
that allows mapping from Riemannian manifolds to Euclidean spaces.
We then describe a classification algorithm that uses WRLPP.


\subsection{Riemannian manifolds}
\label{sec:manifolds}

A manifold can be considered as a continuous surface lying in a higher dimensional Euclidean space.
Formally, a manifold is a topological space which is locally similar to an Euclidean space~\cite{TuzelEtAl2008}.
Intuitively, the tangent space {\eqsize $T_\mat{X}$} is the plane tangent to the surface of the manifold at point~{\eqsize $\mat{X}$}.

A point {\eqsize $\mat{Y}$} on the manifold can be mapped to a vector in the tangent space {\eqsize $T_{\mat{X}}$}
using the logarithm map operator {\eqsize $\log_{\mat{X}}$}.
For~{\eqsize $Sym_d^+$} the logarithm map is defined as:

\vspace{-1ex}
\eqsize
\begin{equation}
    \log_{\mat{X}}\left( \mat{Y} \right) = \mat{X}^{\frac{1}{2}}\log \left(\mat{X}^{-\frac{1}{2}} \mat{Y}\mat{X}^{-\frac{1}{2}} \right) \mat{X}^{\frac{1}{2}}
    \label{eqn:sym_log_map}
\end{equation}
\normalsize

\noindent
where {\eqsize $\log\left(\cdot\right)$} is the matrix logarithm operator.
Given the eigenvalue decomposition of a symmetric matrix,
\mbox{\eqsize $\mat{\Sigma} = \mat{U}\mat{D}\mat{U}^T$},
the matrix logarithm can be computed via:

\vspace{-1ex}
\eqsize
\begin{equation}
    \log(\mat{\Sigma}) = \mat{U} \log(\mat{D}) \mat{U}^T
\end{equation}
\normalsize

\noindent
where {\small $\log(\mat{D})$} is a diagonal matrix,
with each diagonal element equal to the logarithm of the corresponding element in {\small $\mat{D}$}.

The minimum length curve connecting two points on the manifold is called the geodesic,
and the distance between two points is given by the length of this curve.
Geodesics are related to the tangents in the tangent space.
For {\eqsize $Sym_d^+$}, the distance between two points on the manifold can be found via:

\noindent
\eqsize
\begin{equation}
  d^2(\mat{X},\mat{Y}) = \operatorname{trace}\left\{ \log^2 \left( \mat{X}^{-\frac{1}{2}} \mat{Y} \mat{X}^{-\frac{1}{2}} \right) \right\}
  \label{eq:geodesic}
\end{equation}
\normalsize

\subsection{Weighted RLPP}
\label{sec:wrlpp}



The usual approach for classification on manifolds
is to first map the points into an appropriate Euclidean representation~\cite{TuzelEtAl2008}
and then use traditional machine learning methods.
Points in the manifold can be mapped into a fixed tangent space (such as $T_I$ where $I$ is the identity matrix) \cite{GuoEtAl2010}.
Since distances in the manifold are only locally preserved in the tangent space,
better results can be achieved by considering the tangent space at the Karcher mean,
the point which minimises the distances among the samples, as shown in~\cite{TuzelEtAl2008}.
Improved results have been obtained by considering multiple tangent spaces~\cite{Lui2010,SaninEtAl2012}.
A more complex approach involves using training data to create a mapping that tries to preserve
the relations between points, such as the RLPP approach~\cite{HarandiEtAl2012}.


RLPP is based on Laplacian eigenmaps~\cite{BelkinAndNiyogi2003}.
Given {\eqsize $N$} training points
\mbox{\eqsize $\mathbb{X}= \left\{\mat{X}_1,\mat{X}_2, \cdots, \mat{X}_N\right\}$}
from the underlying Riemannian manifold~{\eqsize $\mathcal{M}$},
the local geometrical structure of~{\eqsize $\mathcal{M}$} can be modelled by building an adjacency graph~{\eqsize $\mat{G}$}.
The simplest form of {\eqsize $\mat{G}$} is a binary graph obtained based on the nearest neighbour properties of Riemannian points:
two nodes are connected by an edge if one node is among the $k$ nearest neighbours of the other node.
From the adjacency graph {\eqsize $\mat{G}$} we can find the degree and Laplacian matrices, respectively:

\vspace{-1ex}
\noindent
\eqsize
\begin{eqnarray}
  \mat{D}(i,i) &=& \sum\nolimits_k \mat{G}(i,k)\\
  \mat{L}      &=& \mat{D} - \mat{G}
\end{eqnarray}
\normalsize

\noindent
where the degree
matrix {\eqsize $\mat{D}$} is a diagonal matrix of size \mbox{\eqsize $N \times N$},
with diagonal entries indicating the the number of edges of each node in the adjacency graph.

RLPP also uses a heat pseudo-kernel matrix {\eqsize $\mathbb{K}$},
with the {\eqsize $(i,j)$}-th element constructed via:

\noindent
\eqsize
\begin{equation}
  \mathbb{K}(i,j) = k(\mat{X}_i,\mat{X}_j) = \exp \left\{-\frac{d(\mat{X}_i,\mat{X}_j)}{\sigma} \right\}
  \label{eq:kernel_matrix}
\end{equation}
\normalsize

\noindent
where {\eqsize $d(\cdot, \cdot)$} is the geodesic distance defined in~\eqref{eq:geodesic}.

The final mapping can be found through the following generalised eigenvalue problem~\cite{HarandiEtAl2012}:

\vspace{-1ex}
\eqsize
\begin{equation}
    \mathbb{K} \mat{L} \mathbb{K}^T \mathbb{A} = \lambda \mathbb{K} \mat{D} \mathbb{K}^T \mathbb{A}
\end{equation}
\normalsize

\noindent
where the eigenvectors with the {\eqsize $r$} smallest eigenvalues form the projection matrix {\eqsize $\mathbb{A}$}.

The number of possible \cov~descriptors inside a sample video is very large.
As such, we elected to use boosting to search for a subset of the best descriptors for classification.
We could use the original RLPP mapping approach to map the matrices as vectors at each boosting iteration.
However, as shown in~\cite{TuzelEtAl2008}, the sample weights can be used to generate a mapping
which is more appropriate for the critical training samples.
Therefore, we propose a modified projection, specifically designed to be used during boosting,
which uses sample weights to generate the final mapping.
We refer to this approach as weighted Riemannian locality preserving projection (\wrlpp).

In the modified projection,
the adjacency graph~{\eqsize $\mat{G}$} is replaced 
with a weighted adjacency graph~{\eqsize $\widehat{\mat{G}}$},
defined as:

\vspace{-1ex}
\eqsize
\begin{equation}
  \widehat{\mat{G}} = \mat{W} \mat{G} \mat{W}
\end{equation}
\normalsize

\noindent
where {\eqsize $\mat{W}$} is a diagonal matrix with diagonal values
that correspond to the vector of sample weights {\eqsize $[ w_1, w_2, \ldots, w_N ]$}.
Using the weighted adjacency graph, edges involving critical samples
(ie.~samples with higher weights) become more important and their geometrical structure is better preserved.
The modified projection approach is detailed in~\alg{alg:wrlpp}.

\begin{algorithm}
  \footnotesize
  \raggedright
  \caption{{\bf :} {\footnotesize obtaining weighted RLPP}}
  \label{alg:wrlpp}
  
  \textbf{Input:} Training samples (covariance matrices), labels and weights\\
                  ~~~~~~~~~~~ $\{(\mat{X}_i,y_i,w_i)\}_{i=1}^{N}$, $\mat{X}_i\in\mathcal{M}$\\
  
  \begin{itemize} \leftskip-1em
    \item
      Create Riemannian pseudo-kernel matrix:\\
      $\mathbb{K}(i,j) = \exp \left\{-\frac{d(X_i,X_j)}{\sigma} \right\}$ ~~~ using \eqref{eq:geodesic} as $d(\cdot,\cdot)$
               
    \item
      Construct weighted adjacency graph:\\
      $\widehat{\mat{G}}(i,j) = \left\{ \begin{array}{l l}
                                   w_i \cdot w_j & \text{if} ~ y_i = y_j ~ \text{and} ~ \mat{X}_j ~ \text{is among the}\\
                                                 & k ~ \text{nearest neighbours of} ~ \mat{X}_i ~ \text{in} ~ \mathbb{K}.\\
                                   0             & \text{otherwise}
                                 \end{array} \right.$
    \item Obtain the weighted degree $N \times N$ diagonal matrix:\\
               $\widehat{\mat{D}}(i,i) = \sum_k \widehat{\mat{G}}(i,k)$
    \item Calculate the weighted Laplacian matrix:\\
          $\widehat{\mat{L}} = \widehat{\mat{D}} - \widehat{\mat{G}}$
    \item The eigenvectors with the $r$ smallest eigenvalues of the Rayleigh quotient $\frac{\mathbb{K} \widehat{\mat{D}} \mathbb{K}^T}{\mathbb{K}
          \widehat{\mat{L}} \mathbb{K}^T}$ form the projection matrix $\mathbb{A}$.
  \end{itemize}
  
  \textbf{Output:} Projection model $\lambda = \{\mathbb{A}, \{\mat{X}_i\}_{i=1}^{N}\}$
\end{algorithm}

Once the the projection matrix {\eqsize $\mathbb{A}$} has been obtained,
a given point {\eqsize $\mat{C}$} (a {\cov} matrix) on the manifold can then be mapped to Euclidean space via:

\vspace{-1ex}
\eqsize
\begin{equation}
  \operatorname{WRLPP}(\mat{C}) = \mathbb{A}^T \mat{K}_C
  \label{eq:wrlpp}
\end{equation}
\normalsize

\noindent
where \mbox{\eqsize $\mat{K}_C = \left[ k(\mat{X}_1, \mat{C}),~ k(\mat{X}_2, \mat{C}),~ \cdots\hspace{-0.25ex},~ k(\mat{X}_N, \mat{C}) \right]^T$},
with {\eqsize $k(\cdot, \cdot)$} defined in \eqref{eq:kernel_matrix},
and {\eqsize $\{\mat{X}_i\}_{i=1}^{N}$} representing the training points.

%

%
%
%

\subsection{Classification}
\label{sec:classifier}

As mentioned in the preceding section,
we have chosen to use boosting to find a subset of the best descriptors for classification,
as the number of possible \cov~descriptors inside a sample video is large.
For simplicity, we used a combination of one-vs-one LogitBoost classifiers~\cite{FriedmanEtAl2000}
to achieve multiclass classification.

We start with a brief description of binary LogitBoost classification, with class labels {\eqsize $y_i \in \{0,1\}$}.
The probability of sample {\eqsize $\vect{x}$} belonging to class {\eqsize $1$} is represented by:

\vspace{-1ex}
\eqsize
\begin{equation}
  p(\vect{x})
  =
  \frac
  {\exp\{F(\vect{x})\} }
  { \exp\{F(\vect{x})\} + \exp\{-F(\vect{x})\}}
\end{equation}
\normalsize

\noindent
where {\eqsize $F(\vect{x}) = \frac{1}{2} \sum\nolimits_{m=1}^M g_m(\vect{x})$},
with {\eqsize $g(\vect{x})$} representing a weak learner.

The LogitBoost algorithm learns a set of {\eqsize $M$} weak learners
by minimising the negative binomial log likelihood of the data.
A weighted least squares regression {\eqsize $g_m(\vect{x})$} of training points {\eqsize $\vect{x}_i \in
\mathbb{R}^d$} is fitted to response values {\eqsize $z_i \in \mathbb{R}$}, with weights {\eqsize $w_i$},
where

\noindent
\eqsize
\begin{align}
  w_i = p(\vect{x}_i)(1-p(\vect{x}_i))\\
  z_i = \frac{y_i-p(\vect{x}_i)}{p(\vect{x}_i)(1-p(\vect{x}_i))}
\end{align}
\normalsize

As we are using \cov~descriptors (covariance matrices) as input data,
we adapt the weak learners {\eqsize $g_m(\vect{\cdot})$} to use the projected descriptors.
In other words, {\eqsize $g_m(\vect{x})$} is replaced with {\eqsize $g_m(\operatorname{WRLPP}(\mat{X}))$},
with {\eqsize $\mat{X}$} representing a covariance matrix.

For every unique pair of classes, we train a one-vs-one LogitBoost classifier as follows.
Only the samples belonging to the pair of classes are used for training the binary classifier.
One class is selected to be the positive class and the other as the negative class.
For each boosting iteration, we search for the region whose {\cov} descriptor best separates positive from negative samples. The descriptor is
calculated for all the training samples and mapped to vector space with \wrlpp, using the sample weights calculated for the current boosting
iteration. Once in vector space, we fit a linear regression and use it as the weak LogitBoost classifier.

To prevent overfitting, the number of weak classifiers on each one-vs-one classifier
is controlled by a probability margin between the last accepted positive and the last rejected negative.
Both margin samples are determined by the target detection rate ({\eqsize $dr$}) and the target false positive rejection rate ({\eqsize $rr$}).
The final multiclass classifier is a set of one-vs-one classifiers.
Each one-vs-one classifier {\eqsize $C_{<k,l>}$}, where {\eqsize $k$} and {\eqsize $l$} are the labels of its two classes, has a positive class
{\eqsize $y_{<k,l>}$} and a threshold {\eqsize $\tau_{<k,l>}$}. The positive class is the label of the class deemed to be positive and the threshold
is found via boosting.
\alg{alg:logitboost} summarises the training process.

\begin{algorithm}
  \footnotesize
  \raggedright
  \caption{{\bf :} {\footnotesize Boosting with WRLPP}}
  \label{alg:logitboost}
  
  \textbf{Input:} Training videos with labels $\{(V_i,y_i)\}_{i=1}^{N}$ belonging to $N_c$ classes\\
  
  \begin{itemize} \leftskip-1em
    \item For each unique pair of class labels $<k,l>$ train the one-vs-one classifier $C_{<k,l>}$
      \begin{itemize} \leftskip-1em
        \item Let $k$ be the positive class label and restrict the training set to
              $\{(V_j,y_j)\}_{j=1...N|y_j \in \{k,l\}}$
        \item Let either $k$ or $l$ be the positive label $y_{<k,l>}$
        \item Create binary labels $y'_j \leftarrow (y_j = y_{<k,l>})$
        \item Start with $w_j=1/N$, $F(V)=0$, $p(V_j)=\frac{1}{2}$, $m=1$
        \item Repeat while $p(V_p) - p(V_n) < margin$
          \begin{itemize} \leftskip-1em
            \item Compute the response values and weights $z_j=\frac{y'_j-p(V_j)}{p(V_j)(1-p(V_j))}$, $w_j=p(V_j)(1-p(V_j))$
            \item For each spatio-temporal window $R_s$
              \begin{itemize} \leftskip-2em
                \item Construct the descriptors $X_{j,s}=\widehat{\cov}_{j,R_s}$
                \item From $\{(X_{j,s},y_j,w_j)\}_{j=1}^{N}$ obtain the projection model $\wrlpp_s$ using \alg{alg:wrlpp}
                \item Map the data points $\vect{x}_{j,s} \mbox{=} \wrlpp_s(X_{j,s})$ using~\eqref{eq:wrlpp}
                \item Fit function $g_s(\vect{x})$ by weighted least-squares regression of $z_j$ to $\vect{x}_{j,s}$ using weights $w_j$
              \end{itemize}
            \item Update $F(V) \leftarrow F(V) + \frac{1}{2}f_m(V)$, where $f_m$ is the best classifier among $\{f_s\}$ which
                  minimises the negative binomial log-likelihood $-\sum_{j=1}^N{[y'_j\log(p(\vect{x}_j))+(1-y'_j)\log(1-p(\vect{x}_j))]}$
            \item Update $p(V) \leftarrow \frac{e^{F(V)}}{e^{F(V)}+e^{-F(V)}}$
            \item Sort positive and negative samples according to descending probabilities and find samples at the decision boundaries
                  $V_p=(dr \cdot N_p)$-th $V^+$, $V_n=(rr \cdot N_n)$-th $V^-$, where $dr$ and $rr$ are the desired detection and false positive
                  rejection rates
            \item $m \leftarrow m + 1$
          \end{itemize}
        \item Store $C_{<k,l>}=\{(R_m,\wrlpp_m,g_m)\}_{m=1}^{M}$, threshold $\tau_{<k,l>}=F(V_n)$ and positive label $y_{<k,l>}$
      \end{itemize}
  \end{itemize}
  
  \textbf{Output:} A set of $\frac{N_c(N_c-1)}{2}$ one-vs-one classifiers
\end{algorithm}

A sample video {\eqsize $V$} is classified as follows.
Given a one-vs-one classifier {\eqsize $C_{<k,l>}$},
the probability of a sample video {\eqsize $V$} belonging to the positive class {\eqsize $y_{<k,l>}$} is evaluated using:

\vspace{-2ex}
\eqsize
\begin{equation}
\label{eq:logitboost}
  C_{<k,l>}(V) = \sum\limits_{m=1}^M g_m \left( \operatorname{WRLPP}_m(\widehat{\cov}_{R_m}) \right) - \tau_{<k,l>}
\end{equation}
\normalsize

After evaluating {\eqsize $V$} with all the one-vs-one classifiers in the set,
the sample is labelled as the class {\small $a$} which maximises:

\vspace{-2ex}
\eqsize
\begin{equation}
  C(V) = \arg \max_a \sum\nolimits_{i \neq a} C_{<a,i>}(V) \operatorname{sign_a}(C_{<a,i>}(V))
\end{equation}
\normalsize

\noindent
where {\eqsize $\operatorname{sign_a}(C_{<a,i>}(V))$} is {\eqsize $\operatorname{sign}(C_{<a,i>}(V))$}
if {\small $a$} is the positive class {\small $y_{<a,i>}$},
or {\small $1 - \operatorname{sign}(C_{<a,i>}(V))$} otherwise.
In other words, {\eqsize $V$} is labelled as the class with greater probability sum,
selecting all the one-vs-one classifiers that evaluate to that class.
\section{Experiments}
\label{sec:experiments}

We compared the performance of the proposed algorithm against 
baseline approaches as well as several state-of-the-art methods.
We used three benchmark datasets, with an overview of the datasets shown in \tab{tab:datasets}.

In the following subsections,
we first present an evaluation of several Riemannian to Euclidean space mapping approaches,
justifying the use of the weighted RLPP.
We then follow with experiments showing the performance on sport actions,
facial expressions and hand gestures.

Unless otherwise stated, no pre-processing was performed in the input sequences
and all the recognition results were obtained using 5-fold cross validation to divide the samples into training and testing sets.

In all cases we used the following parameters:
0.95 detection rate, 0.95 false positive rejection rate, 0.5 margin.
Furthermore, since the search space of spatio-temporal windows is very large,
we restricted the minimum size of the windows,
as well as the minimum increment on location and size of the windows, to {\small $\frac{1}{8}$} of the frame size.

\begin{table}[!tb]
  \centering
  \footnotesize
  \begin{tabular}{lccc}
    \toprule
    \hspace{-2ex}
    {\bf Dataset}
    &\begin{minipage}{14ex}
      \centering
      \includegraphics[width=\columnwidth]{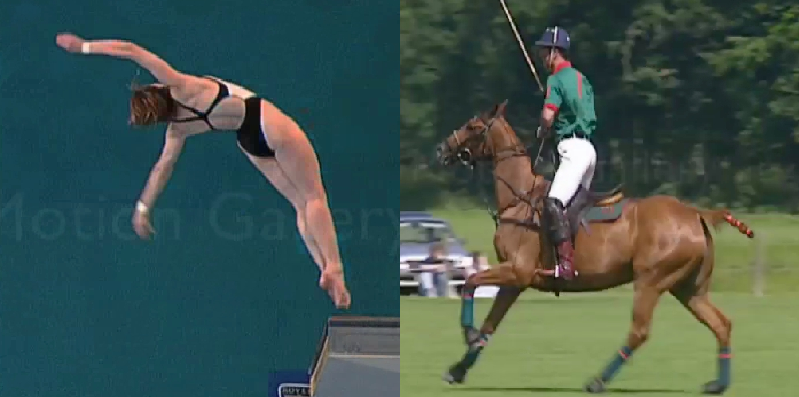}
      {\bf UCF~\cite{RodriguezEtAl2008}}
    \end{minipage}
    &\begin{minipage}{14ex}
      \centering
      \includegraphics[width=\columnwidth]{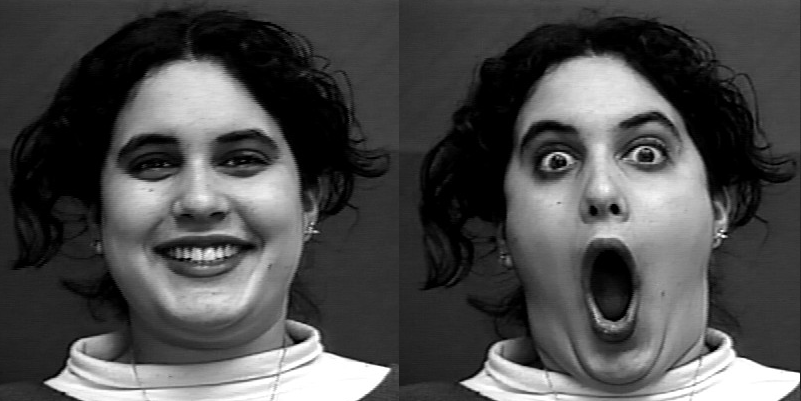}
      {\bf CK+~\cite{LuceyEtAl2010}}
    \end{minipage}
    &\begin{minipage}{14ex}
      \centering
      \includegraphics[width=\columnwidth]{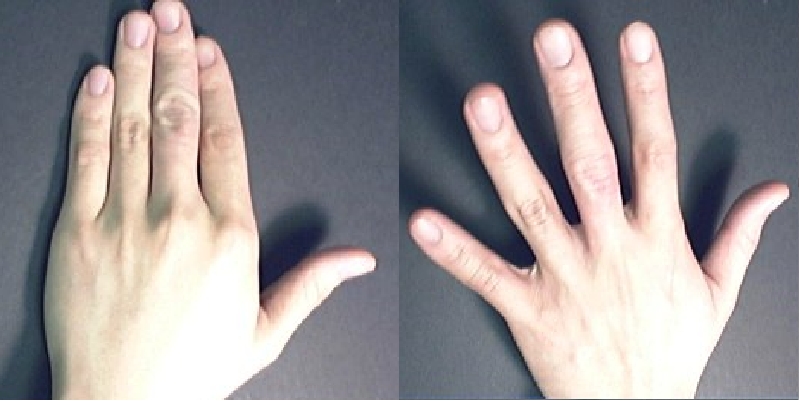}
      \mbox{\bf Cambridge~\cite{KimAndCipolla2009}}
    \end{minipage}\\
    \midrule[\heavyrulewidth]
    \hspace{-1ex}\bf{Type}          &sports           &facial expressions &hand gestures \\
    \hspace{-1ex}\bf{Classes}       &10               &7                  &9\\
    \hspace{-1ex}\bf{Subjects}      &---              &123                &2\\
    \hspace{-1ex}\bf{Scenarios}     &---              &---                &5\\
    \hspace{-1ex}\bf{Video samples}\hspace{-10ex}
                       &150              &593                &900\\
    \hspace{-1ex}\bf{Resolution}    &variable         &$640 \times 480$   &$320 \times 240$\\
    \bottomrule
  \end{tabular}
  
  ~
  
  \caption
    {
    \small
    Overview of the datasets used in the experiments.
    }
  \label{tab:datasets}
\end{table}

\subsection{Comparison of mapping approaches}
\vspace{-0.5ex}

\begin{figure}[!tb]
  \centering
  \includegraphics[width=0.9\columnwidth]{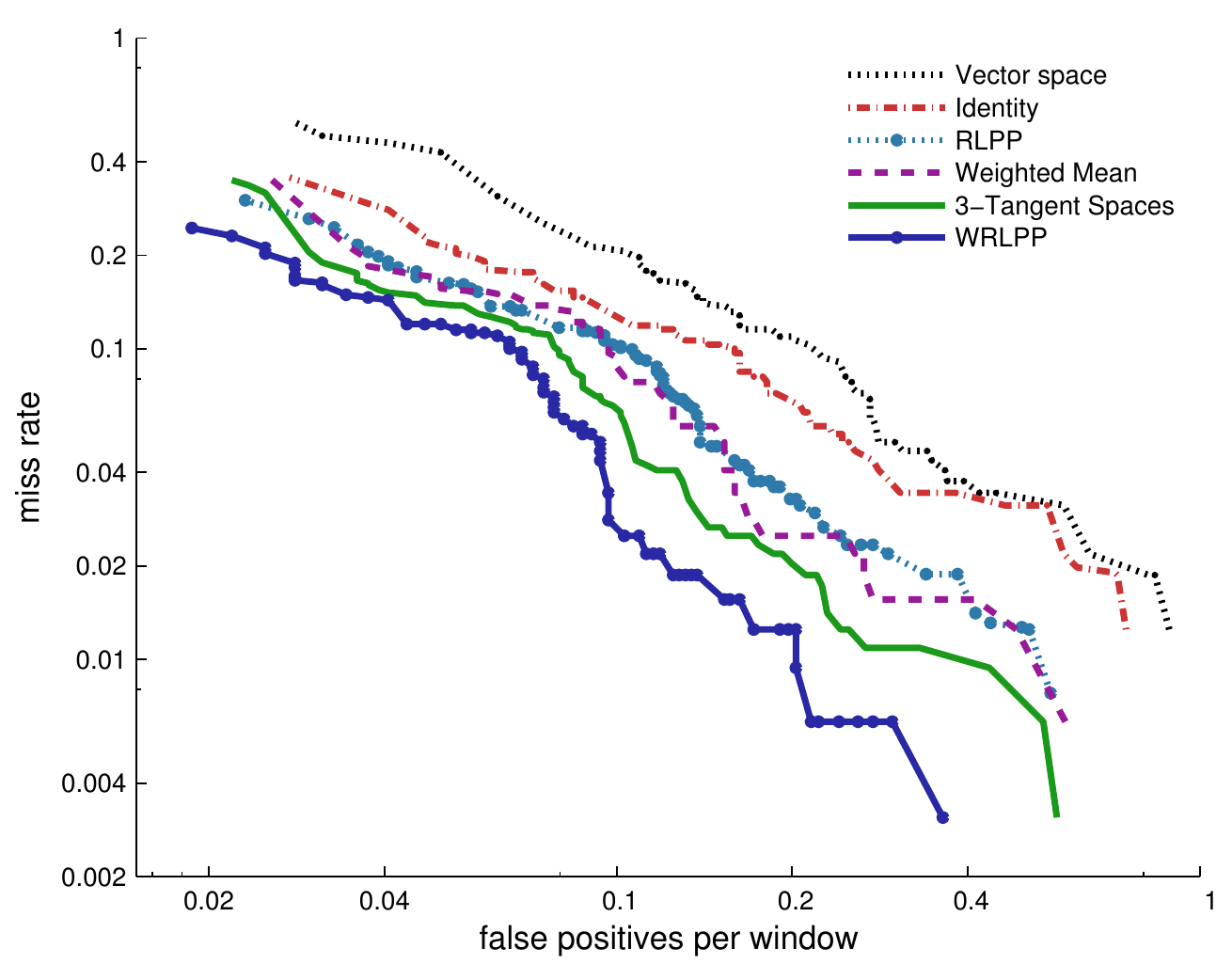}
  \caption
    {
    \small
    Performance comparison of various {\eqsize $Sym_d^+~\mapsto~\mathbb{R}$} mapping approaches,
    used within the classifier framework described in Section~\ref{sec:classifier}.
    }
  \label{fig:mappings}
\end{figure}

In \fig{fig:mappings}, we compare the following six Riemannian to Euclidean space mapping ({\eqsize $Sym_d^+ \mapsto \mathbb{R}$})
approaches which can be used during boosting:
{\bf (i)}~no mapping (ie., using a vectorised representation of the upper-triangle of the covariance matrix),
{\bf (ii)}~projection to a fixed tangent space~\cite{GuoEtAl2010},
{\bf (iii)}~projection to the weighted Karcher mean of the samples~\cite{TuzelEtAl2008},
{\bf (iv)}~projection using k-tangent spaces~\cite{SaninEtAl2012},
{\bf (v)}~mapping the points with the original RLPP method~\cite{HarandiEtAl2012},
and
{\bf (vi)}~mapping the points with the proposed \wrlpp~approach.

Since the mapping approach affects individual binary classifiers,
we show results per classifier with detection error trade-off curves.
We chose the one-vs-one classifiers between conflicting class pairs
(where samples of one class are misclassified as the other class)
on the Cambridge hand gesture recognition dataset
(which is described in Section~\ref{sec:camb_hand_gesture}). Each point on the curve represents the average of all the chosen classifiers.
The curves were obtained by varying the classification threshold {\eqsize $\tau$} in \alg{alg:logitboost}.

With the exception of the original RLPP method, incrementally better results are obtained by using the mapping approaches in the mentioned order,
as they provide increasingly better vector representations of the manifold space.
Although RLPP is designed to provide a better representation compared to tangent-based approaches,
it appears not to be appropriate for boosting as it does not take into account the sample weights of critical training points.
The proposed WRLPP method addresses this problem, resulting in the best overall performance.

\subsection{UCF sport dataset}
\vspace{-0.5ex}

The UCF sport action dataset~\cite{RodriguezEtAl2008} consists of ten categories of human actions,
containing videos with non-uniform backgrounds where both the camera and the subject might be moving.
We use the regions of interest provided with the dataset.

We compared the \cov~approach against the following methods:
HOG3D~\cite{WangEtAl2009},
hierarchy of discriminative space-time neighbourhood features (HDN)~\cite{KovashkaAndGrauman2010},
and augmented features in conjunction with multiple kernel learning (AFMKL)~\cite{WuEtAl2011}.
HOG3D is the extension of histogram of oriented gradient descriptor~\cite{LaptevEtAl2008} to the spatio-temporal case.
HDN learns shapes of space-time feature neighbourhoods that are most discriminative for a given action category.
The idea is to form new features composed of the neighbourhoods around the interest points in a video.
AFMKL exploits appearance distribution features and spatio-temporal context features in a learning scheme for action recognition.
As shown in Table~\ref{tab:ucf_rates}, the proposed \cov-based approach achieves the highest accuracy.

\begin{table}[!tb]
  \centering
  \footnotesize
  \begin{tabular}{lc}
    \toprule
    \bf{Method}                       &\bf{Performance}\\
    \midrule[\heavyrulewidth]
    HOG3D~\cite{WangEtAl2009}         &85.60\%\\
    HDN~\cite{KovashkaAndGrauman2010} &87.27\%\\
    AFMKL~\cite{WuEtAl2011}           &91.30\%\\
    \bf{\cov}                         &\bf{93.91\%}\\
    \bottomrule
  \end{tabular}
  
  ~
  
  \caption
    {
    \small
    Average recognition rate on the UCF dataset~\cite{RodriguezEtAl2008}.
    }
  \label{tab:ucf_rates}
\end{table}

\subsection{CK+ facial expression dataset}

The extended Cohn-Kanade (CK+) facial expression database~\cite{LuceyEtAl2010} contains 593 sequences from 123 subjects.
We used the sequences with validated emotion labels, among 7 possible emotions.
The image sequences vary in duration (i.e. 10 to 60 frames)
and incorporate the onset (which is also the neutral frame) to peak formation of the facial expressions.

We compared the \cov~approach against
active appearance models (AAM),
constrained local models (CLM)~\cite{ChewEtAl2011},
and temporal modelling of shapes (TMS)~\cite{JainEtAl2011}.
AAM is the baseline approach included with the dataset.
It uses active appearance models to track the faces and extract the features,
and then uses support vector machines (SVM) to classify the facial expressions.
The CLM approach is an improvement on AAM, designed for better generalisation to unseen objects.
The TMS approach uses latent-dynamic conditional random fields to model temporal variations within shapes.

We show the performance per emotion in \tab{tab:ck_rates},
in line with existing literature.
The proposed \cov~approach achieves the highest average recognition accuracy of $92.3\%$ (averaged over the 7 classes).
The next best method (TMS) obtained an average accuracy of $87.92\%$.

\begin{table}[!tb]
  \centering
  \footnotesize
  \begin{tabular}{lccccccc}
    \toprule
    \bf{Method}              &\rh{angry} &\rh{contempt} &\rh{disgust} &\rh{fear} &\rh{happy} &\rh{sadness} &\rh{surprise}\\
    \midrule[\heavyrulewidth]
    AAM~\cite{LuceyEtAl2010} &75.0       &84.4          &94.7         &65.2      &\bf{100}   &68.0      &96.0\\
    CLM~\cite{ChewEtAl2011}  &70.1       &52.4          &92.5         &72.1      &94.2       &45.9      &93.6\\
    TMS~\cite{JainEtAl2011}  &76.7       &---           &81.5         &94.4      &98.6       &\bf{77.2} &99.1\\
    \bf{\cov}                &\bf{94.4}  &\bf{100}      &\bf{95.5}    &\bf{90.0} &96.2       &70.0      &\bf{100}\\
    \bottomrule
  \end{tabular}
  
  ~
  
  \caption
    {
    \small
    Recognition rate (in \%) on the CK+ dataset~\cite{LuceyEtAl2010}.
    }
  \label{tab:ck_rates}
\end{table}

\subsection{Cambridge hand gesture dataset}
\label{sec:camb_hand_gesture}

The Cambridge hand-gesture dataset~\cite{KimAndCipolla2009} consists of 900 image sequences of 9 gesture classes.
Each class has 100 image sequences performed by 2 subjects,
captured under 5 illuminations and 10 arbitrary motions.
The 9 classes are defined by three  primitive hand shapes and three primitive motions.
Each sequence was recorded with a fixed camera having roughly isolated gestures in space and time.
We followed the test protocol defined in~\cite{KimAndCipolla2009}.
Sequences with normal illumination were considered for training while tests were performed on the remaining sequences.

The proposed method was compared against tensor canonical correlation analysis (TCCA)~\cite{KimAndCipolla2009}, product manifolds
(PM)~\cite{LuiEtAl2010} and tangent bundles (TB)~\cite{Lui2010}. TCCA is the extension of canonical correlation analysis to multiway data arrays or
tensors. Canonical correlation analysis and principal angles are standard methods for measuring the similarity between subspaces.
In the PM method a tensor is characterised as a point on a product manifold and  classification is performed on this space. The product manifold is
created by applying a modified high order singular value decomposition on the tensors and interpreting each factorised space as a Grassmann manifold.
In the TB method, video data is represented as a third order tensor and factorised using high order singular value decomposition, where each factor is
projected onto a tangent space and the intrinsic distance is computed from a tangent bundle for action classification.

We report the recognition rates for the four test sets in Table~\ref{tab:hands_rates},
where the proposed \cov-based approach obtains the highest performance.

\begin{table}[!tb]
  \centering
  \footnotesize
  \begin{tabular}{lccccc}
    \toprule
    \bf{Method}                   &\bf{Set1} &\bf{Set2} &\bf{Set3} &\bf{Set4} &\bf{Overall}\\
    \midrule[\heavyrulewidth]
    TCCA~\cite{KimAndCipolla2009} &81\%      &81\%      &78\%      &86\%      &82\% ($\pm$3.5)\\
    PM~\cite{LuiEtAl2010}         &89\%      &86\%      &89\%      &87\%      &88\% ($\pm$2.1)\\
    TB~\cite{Lui2010}             &\bf{93\%} &88\%      &90\%      &91\%      &91\% ($\pm$2.4)\\
    \bf{\cov}                     &92\%      &\bf{94\%} &\bf{94\%} &\bf{93\%} &\bf{93\% ($\pm$1.1)}\\
    \bottomrule
  \end{tabular}
  
  ~
  
  \caption
    {
    \small
    Average recognition rate on the Cambridge dataset~\cite{KimAndCipolla2009}.
    }
  \label{tab:hands_rates}
\end{table}

\section{Conclusion}
\label{sec:discussion}

In this paper, we first extended the flat covariance descriptors proposed in~\cite{TuzelEtAl2008}
to spatio-temporal covariance descriptors termed \cov,
and then showed how they can be computed quickly through the use of integral video representations. 

The proposed \cov~descriptors belong to the group of symmetric positive definite matrices,
which can be formulated as a connected Riemannian manifold.
Prior to classification, points on a manifold are generally mapped to an Euclidean space,
through a technique such as Riemannian locality preserving projection (RLPP)~\cite{HarandiEtAl2012}.

The \cov~descriptors are extracted from spatio-temporal windows inside sample videos,
with the number of possible windows being very large.
We used a boosting approach to find a subset which is the most useful for classification.
In order to take into account the weights of the training samples,
we further proposed to extend RLPP by incorporating weighting during the projection.
The weighted projection (termed WRLPP) leads to a better representation of the
neighbourhoods around the most critical training samples during each boosting iteration.

Combining the proposed \cov~descriptors with the classification approach based on WRLPP boosting
leads to a state-of-the-art method for action and gesture recognition.
The proposed \cov-based method performs better than several recent approaches 
on three benchmark datasets for action and gesture recognition.
The method is robust and does not require additional processing of the videos,
such as foreground detection, interest-point detection or tracking.
To our knowledge, this is the first approach proving to be equally suitable (ie., $>90\%$ recognition accuracy)
for both action and gesture recognition.

Further avenues of research include adapting the method for related tasks,
such as anomaly detection in surveillance videos~\cite{reddy_cvprw_2011},
where there is often a shortage of positive examples.


\balance
\renewcommand{\baselinestretch}{0.983}\small\footnotesize
\footnotesize
\bibliographystyle{ieee}
\bibliography{references}

\end{document}